\documentclass[runningheads]{llncs}
\usepackage{graphicx}
\usepackage{subfigure}
\usepackage{hyperref}
\usepackage{multirow}
\usepackage{array}
\usepackage[numbers]{natbib}

\usepackage{url}

\usepackage{breakurl}

\usepackage[textsize=scriptsize,backgroundcolor=green]{todonotes}

\newcommand\f[1]{{\bf #1}}
\newcommand\waseem[0]{W\&H }

\begin{document}
\title{You Are What You Tweet: Profiling Users by Past Tweets to Improve Hate Speech Detection}
\titlerunning{Profiling Users by Past Tweets to Improve Hate Speech Detection}

\author{
Prateek Chaudhry and Matthew Lease}
\institute{ 
  University of Texas at Austin \\
  \email{prateek.chaudhry@gmail.com, ml@utexas.edu}}

\maketitle

\begin{abstract}
Hate speech detection research has predominantly focused on purely content-based methods, without exploiting other contextual data. We briefly critique pros and cons of this task formulation. We then investigate profiling users by their past utterances as an informative prior to better predict whether new utterances constitute hate speech. To evaluate this, we augment three Twitter hate speech datasets with additional timeline data, then embed this additional context into a strong baseline model. Promising results suggest merit for further investigation. 

\end{abstract}

\section{Introduction}
\label{sec:intro}





%
Online hate speech is a vast and continually growing problem \cite{halevy2020preserving,jurgens2019just, macavaney2019hate,schmidt2017survey}. 
The detection task is most commonly framed as purely content-based: each utterance is classified without any additional context.  
However, history often repeats itself, and ``A user who is known to write hate speech messages may do
so again'' \cite{schmidt2017survey}. Facebook researchers have also recently noted that ``a user or group that has posted violating content in the past may be prone to do so more often in the future'' \cite{halevy2020preserving}. 
This suggests that modeling of user priors may be highly informative and complementary to purely content-based detection models. We return to discussion of ethical considerations in Section \ref{sec:ethics}. 

To investigate user profiling, we augment three Twitter hate speech datasets \cite{arango2019hate,davidson,waseemA} with additional recent {\em timeline} Tweets by each author, then embed this additional context into a strong baseline model \cite{badjatiya2017deep} that was later further refined \cite{arango2019hate}.
Results show strong improvement on one dataset, little benefit on another, and fairly consistent improvement on a third. 
Overall, results across experimental conditions and metrics suggest user  modeling merits further work, though analysis is complicated by differences in annotation schemes and processes, as well as Twitter API limitations and data sharing policies.

\section{Related Work}
\label{sec:related}

%


While most hate speech detection models have been content-oriented and non-contextual, there are notable exceptions. After Hovy \cite{hovy2015demographic} showed modeling value of demographics for classification tasks, Waseem and Hovy \cite{waseemA} applied this to hate speech, inferring gender (by name) and location (by timezone) to classify hate speech with modest benefit: only gender raised hate detection accuracy, and it was statistically significant only when both gender and location were used. A challenge with demographics is that they are sensitive and often not widely available, if  captured at all by the platform. 

Rather than classify hateful utterances, other work has sought to classify hateful users. Assuming ``birds of a feather flock together'', {\em community detection}  \cite{fortunato2010community} has sought to identify hater communities.  Ribeiro et al.~\cite{ribeiro2018characterizing} and Mathew et al.~\cite{mathew2019spread} analyze Twitter's retweet graph to  detect users likely to spread hateful content. Mishra et al.~\cite{mishra2018author} is the only work we are aware of to profile hateful users, via the follower graph, and then use these profiles to improve hate speech detection over baseline models. 

Some hate speech datasets are heavily dominated by a few prolific haters. In Waseem and Hovy~\cite{waseemA}, all racist-labeled tweets came from  9 users. In fact, Arango et al.~\cite{arango2019hate} note that 
a single user generates 96\% of all racist tweets, while another user produces 44\% of all sexist tweets. Arango et al.\ aptly critique that benchmarking on such highly skewed datasets risks overfitting a few individuals rather than learning a broadly applicable model. Similar concerns have been voiced about over-fitting idiosyncrasies of a few annotators \cite{geva2019we}. We agree that datasets should have many diverse examples of the phenomena being modeled (i.e., hate speech examples from many users). However, dataset composition is orthogonal to model design: how can we best model historical context in prediction and fairly evaluate vs.\ context-free models?

Some past studies (cf., \cite{waseemA,xiang2012detecting}) have partitioned train/test data by utterance (i.e., by tweet), rather than by user. The consequence of such experimental design is that any model trained and tested in this manner can be expected to make better predictions on users found in both train and test splits. However, we should not confuse desirable user modeling with questionable experimental design in which user history is only captured haphazardly by whichever users happen to have some tweets in the training data, and the ratio of that user-specific history vs.\ training tweets from other users. Such a scheme does not reflect intentional design or controlled evaluation of user modeling.

An intuitive language-based approach is to model each user by a ``document'' of their past utterances. Such historical data is certainly available to social media companies, and often publicly via API or crawling. The closest work we are aware of on content-based modeling of haters is by Dadvar et al.\ \cite{dadvar2013improving}, though they only considered the user’s past number of obscene words as a modeling feature. 


\section{Datasets}
\label{sec:data}

\begin{figure}%
\centering
\subfigure{%
\label{fig:first}%
     \includegraphics[width=.45\columnwidth]{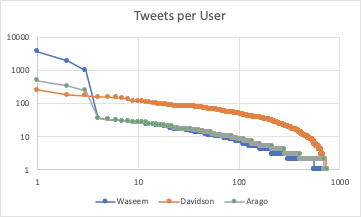}
}
\qquad
\subfigure{%
\label{fig:second}%
     \includegraphics[width=.45\columnwidth]{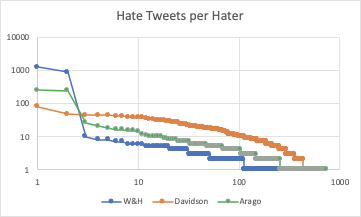}}
\caption{Log-log distribution of tweets per user (top) and hate tweets per user for users producing 1 or more hate tweets (bottom), for all three datasets. The y-axis indicates the number of tweets while the x-axis simply enumerates users in descending order of activity.}
\label{fig:distribution}
\end{figure}

We adopt the same three datasets studied by Arango et al.\  \cite{arango2019hate}; see statistics in {\bf \autoref{table:metrics}}. As is common in other problem domains, each hate speech dataset has various limitations; by evaluating across several datasets, we test across varying data conditions. 

{\bf Waseem and Hovy  \cite{waseemA}} ( \waseem) label 16K tweets for 3 classes: Racism, Sexism and Neither. 
After author labeling, tweets labeled as hate speech were reviewed by a 25-year
old female ``studying gender studies and a nonactivist feminist". Agreement between author and reviewer was $\kappa$= 0.84, with 85\% of  disagreements on sexist labels, and 98\% of these changed to neither.  Tweet IDs and labels were shared, with tweets obtained via the Twitter API. Since tweets and user accounts are often deleted, 15K tweets are found in \cite{arango2019hate}. We find 10K tweets (and only 6 racist ones) over 1,458 users.

{\bf Davidson et al.\  \cite{davidson}} crowd-label 25K tweets for 3 classes: Hate, Offensive and Neither. Majority voting over 3 or more workers was used for label aggregation. 200 tweets were discarded with no majority over the 3 classes. For hate, only 5\% were labeled by 2/3 and only 1.3\% by 3/3. Most were labeled offensive (76\% at 2/3, 53\% at 3/3) and the rest non-offensive (16.6\% at 2/3, 11.8\% at 3/3). We find 18K tweets ($\sim$75\%) from 741 users.

{\bf Arango et al.~\cite{arango2019hate}} identify three issues: 1) \waseem has few haters; 2) those haters are prolific; and 3) train/test splits by utterance mean the same hater often appears in both train and test splits. To address this, they 1) add to \waseem all hate-labeled tweets from DAVIDSON; 2) restrict to at most 250 tweets per user per class; and 3) perform train/test split by user rather than by utterance. Their fused dataset had 7K tweets, collapsing labels to simple binary hate vs.\ non-hate. We find 5.4K over 1746 users. 

{\bf \autoref{fig:distribution}} shows the distribution of tweets per user across datasets. For all users who produce 1 or more hate tweet (W\&H: 443 users, Davidson: 566, Arango: 734), we also plot the number of hate tweets from each such ``hater''.

\section{Profiling Users by Past Tweets}
\label{sec:model}
\label{sec:profile}


As a strong baseline, we adopt Badjatiya et al.\ \cite{badjatiya2017deep}'s 2-phase model, using \cite{arango2019hate}'s corrected version in which word embeddings are derived only from training data.
%
%
An LSTM classifier is first trained to predict the label. Each word is converted to a dense vector representation using a word embedding matrix initialized with pretrained GloVe embeddings \cite{glove}. The final vector from the LSTM is followed by a fully connected layer and a softmax or sigmoid layer for obtaining prediction probability. The network is trained with cross entropy loss with Adam optimizer. Once training of LSTM is done, the first layer is extracted, i.e. word embedding matrix fine tuned on the training set. In the second phase, a tweet is converted to a fixed sized vector by averaging the embedding vectors of its tokens using the trained embeddings from the previous phase. Representing the tweet by this vector, a Gradient Boosted Decision Tree (GBDT) is trained for classification. 

Given a tweet ID, we use the Twitter API to retrieve the author's {\em timeline}\footnote{\url{https://developer.twitter.com/en/docs/tweets/timelines/}}: their latest 20 tweets. This size of 20 reflects an API maximum, but it would be interesting in future work to model longer histories. Given this history, we augment existing public datasets (\autoref{sec:data}) with these timelines. \waseem includes tweet IDs, and Davidson kindly shared their tweet IDs with us upon request. In practice, timeline tweets should precede tweets being classified; here we use existing datasets. By augmenting existing datasets, we can assess the relative benefit with and without user profiling on hate speech datasets that are already familiar to the research community.

We utilize these user profiles in (Arango et al.~\cite{arango2019hate}'s corrected version of) \cite{badjatiya2017deep}'s model as follows. Given a tweet, the author's  timeline is averaged using the same trained word embedding used for tweet representation. Following this, both tweet representation and timeline representation are concatenated and used to train the GBDT classifier. In general, a history-informed prior of the user ought to complement any content-based model of the tweet alone. We expect future work will continue to benefit from exploiting such a prior-informed modeling architecture, irrespective of the specific model.

\section{Experimental Setup}
\label{sec:setup}

We build on Badjatiya et al.\ \cite{badjatiya2017deep}'s and Arango et al.\ \cite{arango2019hate}'s shared source code. We use 10-fold cross validation with default hyperparameters for \cite{badjatiya2017deep}'s model. 
The LSTM word embeddings are initialized with 200 dimensional pretrained GloVe embeddings, and the size of the LSTM representation is 200. We also add dropout of 0.25 and 0.5 after the word embedding layer and the LSTM, respectively. We train the LSTM based architecture for 10 epochs, and then train the GBDT using the final word embeddings.

With nearly all racism tweets in \waseem deleted, we train it as a binary classifier: sexism vs. none. For DAVIDSON, we train a ternary classifier over its three classes. ARANGO fuses \waseem and DAVIDSON label sets by collapsing classes into simple binary classification of hate vs.\ non-hate. 

Tweet deletions make comparison to prior published results more difficult, with only 2/3 of \waseem and 3/4 of DAVIDSON datasets still available, and nearly all \waseem racist tweets deleted.
Arango et al.\ \cite{arango2019hate} note that partitioning train/test by tweet, as in past studies, results in prolific tweeters appearing in both splits. The risk this poses is potentially overfitting to particular users.  Instead, they argue for splitting train/test by user. To be as comprehensive as possible, we report results both ways for completeness.

\section{Results}
\label{sec:results}

\begin{table*}[t]
\centering
\scalebox{0.8}{
\begin{tabular}{|p{1.5cm}|crr||ccc||ccc||ccc||ccc|}
\hline
\multicolumn{4}{|c||}{}&\multicolumn{6}{c||}{Split by Tweet} &\multicolumn{6}{c||}{Split by User} \\
\hline
\multicolumn{4}{|c||}{}&\multicolumn{3}{c||}{Baseline} & \multicolumn{3}{c||}{With Timeline} & \multicolumn{3}{c||}{Baseline} & \multicolumn{3}{c||}{With Timeline}\\
\hline
Dataset & Label & Tweets & Found & P & R & F1 & P & R & F1 & P & R & F1 & P & R & F1 \\
\hline

\multirow{4}{4em}{\waseem} 
       & Racism & 1,851 & 6 & - & - & - & - & - & - & - & - & - & - & - & - \\
       & Sexism & 2,988 & 2,777 & 76.8 & 65.6 & 70.7 & \f{83.7} & \f{80.2} & \f{81.9} & 59.2	& 33.2 & 37.8 & \f{63.7} & \f{35.4}	& \f{40.8}\\
       & Neither & 10,110 & 7,320 & 87.7 & 92.4 & 89.9 & \f{92.6} & \f{94.1} & \f{93.3} & 82.2 &	91.4 &	84.3 & \f{82.6}	& \f{93.9} & \f{86.5}\\
       & Micro Avg &  &  & 85.1 & 85.1 & 85.1 & \f{90.3} & \f{90.3} & \f{90.3} & 76.0 & 76.0 & 76.0 & \f{79.3} & \f{79.3} & \f{79.3} \\
       & Macro Avg &  14,949 & 10,103  & 82.2 & 79.0 & 80.3 & \f{88.2} & \f{87.2} & \f{87.6} & 70.7 & 62.3 & 61.1 & \f{73.1} & \f{64.6} & \f{63.6}\\
\hline

\multirow{4}{4em}{\small DAVID-SON} & Hate & 1,430 & 1,118 & 55.4 & 24.3 & 33.6 & \f{56.4} & \f{24.7} & \f{34.1} & \f{53.6} &	\f{22.1} &	\f{31.0}	&	52.2	& 21.5 &	30.3\\

       & Offensive & 19,190 & 13,931 & \f{91.3} & 95.9 & 93.5 & 91.2 &  95.9 & 93.5 & 91.2 & \f{95.9} &	\f{93.5} &	91.2 &	95.8 &	93.4\\
       & Neither & 4,163 & 2,978 & \f{83.8} & \f{81.5} & \f{82.7} & 83.7 & 81.1 & 82.3 & \f{82.8} & \f{81.3} & \f{82.0} &	82.7 &	81.2 &	81.9\\
       & Micro Avg &  &  & \f{89.1} & \f{89.1} & \f{89.1} & 89.0 & 89.0 & 89.0 & \bf{88.9} & \bf{88.9} & \bf{88.9} & 88.8 & 88.8 & 88.8 \\
       & Macro Avg & 24,783 & 18,027 & 76.8 & \f{67.3} & 69.9 & \f{77.1} & 67.2 & \f{70.0} & \f{75.9} & \f{66.4} & \f{68.8}	& 75.3 & 66.2 & 68.6\\
\hline

\multirow{4}{4em}{\small ARANGO} 
& Hate & 2,920 & 1,988 & 81.0 & 72.4 & 76.4 & \f{85.2} & \f{75.2} & \f{79.8} & \f{80.7} & 63.1 & 68.5 & 79.6 & \f{65.7} & \f{69.8} \\
& None & 4,086 & 3,481 & 85.1 & 90.3 & 87.6 & \f{86.7} & \f{92.5} & \f{89.5} & 81.5 & \f{87.9} & \f{82.8} & \f{82.6} & 86.1 & 82.3 \\
& Micro Avg &  &  & 83.8 & 83.8 & 83.8 & \f{86.1} & \f{86.1} & \f{86.1} & 79.8 & 79.8 & 79.8 & \f{80.1} & \f{80.1} & \f{80.1} \\
& Macro Avg & 7,006 & 5,469 & 83.1 & 81.3 & 82.0 & \f{86.0} & \f{83.8} & \f{84.6} & 81.1 & 75.5 & 75.7 & 81.1 & \f{75.9} & \f{76.1} \\
\hline

\hline
\end{tabular}
} 
\caption{Baseline vs.\ timeline accuracy across datasets and train/test  partitions: by tweet or user. For each (dataset, split, metric) tuple, we bold the higher accuracy between baseline vs.\ timeline.}
\label{table:metrics}
\end{table*}

\begin{table*}[h]
\centering
\scalebox{0.9}{
\begin{tabular}{|m{2cm}|rr||ccc||ccc|}
\hline
\multicolumn{3}{|}{}&\multicolumn{3}{c}{Baseline}&\multicolumn{3}{c|}{With Timeline}\\
\hline
Dataset & Size Timeline & Tweets & P & R & F & P & R & F \\
\hline \multirow{4}{4em}{\waseem} 
            & 0-5 & 58 & \f{78.2} & \f{74.1} & \f{75.8} & 68.8 & 67.4 & 68.0 \\
            & 6-10 & 18 & 47.1 & 47.1 & 47.1 & \f{75.0} & \f{97.1} & \f{81.8} \\
            & 11-15 & 34 & \f{60.0} & \f{54.7} & \f{53.0} & 50.5 & 50.2 & 46.0 \\
            & 16-20 & 9,991 & 82.2 & 79.0 & 80.4 & \f{88.3} & \f{87.4} & \f{87.9} \\
\hline \multirow{4}{4em}{DAVIDSON} 
            & 0-5 & 4,943 & \f{76.6} & \f{66.6} & \f{69.6} & 76.3 & 66.4 & 69.4 \\
            & 6-10 & 22 & 96.5 & 72.2 & 80.4 & NAN & 0 & NAN \\
            & 11-15 & 287 & \f{66.0} & \f{58.8} & \f{61.5} & 58.5 & 54.0 & 55.8 \\
            & 16-20 & 12,775 & 76.9 & 67.5 & 70.1 & \f{77.2} & \f{67.6} & \f{70.3}  \\
\hline
\multirow{4}{4em}{ARANGO} 
            & 0-5 & 59 & \f{76.5} & \f{75.4} & \f{75.9} & 70.0 & 70.0 & 70.0 \\
            & 6-10 & 20 & \f{100.0} & \f{100.0} & \f{100.0} & 97.2 & 83.3 & 88.6 \\
            & 11-15 & 49 & \f{81.2} & \f{80.3} & \f{79.5} & 77.0 & 76.2 & 75.4 \\
            & 16-20 & 5,341 & 83.0 & 81.3 & 82.0 & \f{86.1} & \f{84.0} & \f{84.9} \\
\hline
\multirow{1}{4em}{ARANGO} 
            & 0-5 & 59 & 81.9 & 77.8 & 79.4 & \f{85.3} & \f{78.9} & \f{81.3} \\
\multirow{2}{4em}{(split by user)} 
            & 6-10 & 20 & 97.2 & \f{83.3} & 88.6 & 97.2 & 83.2 & 88.6 \\
            & 11-15 & 49 & 84.9 & 80.8 & 79.2 & 84.9 & 80.8 & 79.2 \\
            & 16-20 & 5,341 & \f{78.3} & 76.1 & 76.9 & 78.2 & \f{77.1} & \f{77.6}\\
\hline
\end{tabular}
} 
\caption{Detection accuracy on user subgroups, based on amount of Twitter timeline available per user (0-20 most recent tweets). Train/test partition of data is by tweet unless noted otherwise. } 
\label{table:metrics_with_timelinelen}
\end{table*}

{\bf \autoref{table:metrics}} compares model performance with and without user timeline history 
across the three datasets and the two train/test data partitions: by tweet vs.\ by user. With regard to testing baseline vs.\ timeline results, splitting by user is cleaner because for any users appearing in the test data, there are no tweets from that user in the training set.

Results show strong improvement on \waseem but little change on DAVIDSON. We do see split-by-tweet shows modest improvement for DAVIDSON on the Hate category, but not on other categories or for split-by-user. Between \waseem and DAVIDSON extremes, we see more modest but fairly consistent improvement on Arango et al.'s dataset, which fuses \waseem and DAVIDSON by adding to \waseem all hate-labeled tweets from DAVIDSON and down-sampling to at most 250 tweets per user per class to reduce skew. Consistent with Arango et al.~\cite{arango2019hate}, we  see much higher results on \waseem and their fused dataset when splitting by tweet rather than by user (Arango et al.\ do not report on DAVIDSON).  

Given the difference in model performance across the datasets, what explains this? \autoref{table:metrics} and \autoref{fig:distribution} show important differences in data scale and distribution across different datasets. The classes being annotated also differ, as does the method of annotation: \waseem uses traditional annotators while DAVIDSON relies on crowd annotators. Such differences highlight several dimensions of ongoing debate in hate speech research surrounding differing approaches to annotation categories and processes \cite{balayn2019designing,fortuna2020toxic,waseemB}. 


We also conducted a further analysis to assess whether the amount of timeline history per user varied significantly across datasets. Results appear in {\bf \autoref{table:metrics_with_timelinelen}}, where we bin users by the count of timeline tweets found: 0-5, 6-10, 11-15, or 16-20. We see that the vast majority of users have 16-20 past timeline tweets, and for this category we see consistent improvement across datasets and train/test split conditions, driving scores. 

Crucially, however, on DAVIDSON we see nearly 5K tweets ($\sim$28\%) from users with only 0-5 timeline tweets available, where we expect small to no improvement from timelines due to lack of history. This holds across datasets when splitting by tweet, including DAVIDSON, which indeed drives down overall results on this dataset. When splitting by user on ARANGO, we do see increased accuracy even with 0-5 timeline tweets, though the sample size is quite small to draw conclusions.  



Further limited qualitative analysis helped give us a sense of examples where profiling a user created an informative prior for correctly classifying tweets that might have been missed otherwise. We explored most prolific tweeters in \waseem for evidence of sexist timeline tweets. As an example, one user made a possibly sexist tweet \textit{``.@USER1 @USER2 when was she good?  i confuse her and ten other women, which is why their pay is lower btw. supply vs. demand.''} which was easier to predict by their earlier timeline tweet \textit{``\#FeminismIsAwful''}. Use of the timeline correctly predicts the `sexist' label, and increases overall detection accuracy for this user from 53\% to 87\%.


\section{Ethical Considerations}
\label{sec:ethics}

While user modeling is common in personalized ranking and recommendation systems (e.g., Google or Netflix), some users may wish to opt-out, and GDPR allows ``the right to be forgotten''. User modeling to prevent fraud and abuse, while equally common in commercial systems, may raise different sorts of ethical and legal questions. We know that social media platforms already do monitor user accounts for terms of service violations, suspending accounts demonstrating repeated abusive behaviors \cite{halevy2020preserving}. Given the vast scale and expense of commercial content moderation, and extensive user histories available, platforms cannot afford to ignore such a strong predictive signal.

However, what challenges may arise? Would a reformed hater have difficulty overcoming his/her past profile a model had learned? One idea would be to decay weight assigned to past tweets by age, or to completely restrict history to a recent window (e.g., our current timeline of past 20 tweets). In fact, the opposite problem is significant on platforms today: a bad actor who is suspended often circumvents this by creating a new account, with a cycle of continuing abusive behaviors by re-entrant bad actors. 


\section{Conclusion}

Producers of hate speech are often repeat offenders, yet we know of no prior work explicitly profiling users by their past tweets to improve hate speech detection accuracy. We collect Twitter timeline data toward this end to augment existing hate speech datasets. Results on several datasets, metrics, and experimental settings are encouraging, but confounds remain. Future work might explore better modeling (e.g., via BERT \cite{devlin2018bert}), collecting more user history, and combining with other profiling approaches \cite{mishra2018author}. Other approaches may explore adding other features extracted from user twitter timeline, such as hate speech tweet frequency, number of hate speech retweets, etc.

\section*{Acknowledgments} 

We thank the reviewers for their valuable feedback, and the many talented workers who provided the hate speech annotations that enable research on it. This research was supported in part by Wipro, the Micron Foundation, and by Good Systems\footnote{\url{http://goodsystems.utexas.edu/}}, a UT Austin Grand Challenge to develop responsible AI technologies. The statements made herein are solely the opinions of the authors and do not reflect the views of the sponsoring agencies.





\bibliographystyle{splncsnat}
\bibliography{paper}
\end{document}